\documentclass[sigconf]{acmart}

\usepackage{booktabs}
\usepackage{url}

\usepackage{times}
\usepackage{xcolor}
\usepackage{soul}
\usepackage[utf8]{inputenc}
\usepackage{caption}

\usepackage{graphics}

\usepackage{bm}
\usepackage{amsmath}
\usepackage{amssymb}
\usepackage{amsfonts}

\usepackage{algorithm}
\usepackage{algpseudocode}

\usepackage{multicol}
\usepackage{subcaption}

\DeclareMathOperator*{\argmin}{arg\,min}
\DeclareMathOperator*{\argmax}{arg\,max}

\newcommand{\figg}{{Figure~}}
\newcommand{\tabb}{{Table~}}
\newcommand{\eqq}{{Eq.~}}

\renewcommand\footnotetextcopyrightpermission[1]{} 
\begin{document}

\title{Internal Model from Observations for Reward Shaping}

\author{Daiki Kimura}
\affiliation{%
  \institution{IBM Research AI}
  \state{Tokyo}
  \country{Japan}
}
\email{daiki@jp.ibm.com}

\author{Subhajit Chaudhury}
\affiliation{%
  \institution{IBM Research AI}
  \state{Tokyo}
  \country{Japan}
}
\email{subhajit@jp.ibm.com}

\author{Ryuki Tachibana}
\affiliation{%
  \institution{IBM Research AI}
  \state{Tokyo}
  \country{Japan}
}
\email{ryuki@jp.ibm.com}

\author{Sakyasingha Dasgupta}
\authornote{This work originated while working at IBM Research}
\affiliation{%
  \institution{Ascent Robotics Inc.}
  \state{Tokyo}
  \country{Japan}
}
\email{sakya@ascent.ai}

\begin{abstract}  
Reinforcement learning methods require careful design involving a reward function to obtain the desired action policy for a given task. In the absence of hand-crafted reward functions, prior work on the topic has proposed several methods for reward estimation by using expert state trajectories and action pairs. However, there are cases where complete or good action information cannot be obtained from expert demonstrations. We propose a novel reinforcement learning method in which the agent learns an internal model of observation on the basis of expert-demonstrated state trajectories to estimate rewards without completely learning the dynamics of the external environment from state-action pairs. The internal model is obtained in the form of a predictive model for the given expert state distribution. During reinforcement learning, the agent predicts the reward as a function of the difference between the actual state and the state predicted by the internal model. We conducted multiple experiments in environments of varying complexity, including the Super Mario Bros and Flappy Bird games. We show our method successfully trains good policies directly from expert game-play videos.
\end{abstract}


\maketitle

\section{Introduction}
Reinforcement learning~(RL)~\cite{Sutton:1998:IRL:551283} enables an agent to learn the desired behavior required to accomplish a given objective, such that the expected return or reward for the agent is maximized over time. Typically, a scalar reward signal is used to guide the agent's behavior so that the agent learns a control policy that maximizes the cumulative scalar reward over trajectories. This type of learning is referred to as \textit{model-free} RL if the agent does not have an apriori model or knowledge of the dynamics of the environment it is acting in. Some notable breakthroughs among the many recent research efforts that incorporate deep models are the deep Q-network (DQN)~\cite{Mnih2015}, which approximated a Q-value function used as a deep neural network and trained agents to play Atari games with discrete control, the deep deterministic policy gradient (DDPG)~\cite{ddpg}, which successfully applied deep RL for continuous control agents, and the trust region policy optimization (TRPO)~\cite{icml2015_schulman15}, which formulated a method for optimizing control policies with guaranteed monotonic improvement.

In most RL methods, it is critical to choose a well-designed reward function to successfully ensure that the agent learns a good action policy for performing the task. Moreover, there are cases in which the reward function is very sparse or may not be directly available. Humans can often imitate the behavior of their instructors and estimate which actions or environmental states are good for the eventual accomplishment of a task without being provided with a continual reward. For example, young adults initially learn how to write letters by imitating demonstrations provided by their teachers or other adults~(experts). Further skills get developed on the basis of exploration around this initial grounding provided by the demonstrations. Taking inspiration from such scenarios, various methods have been proposed, which are collectively known as imitation learning~\cite{ho2016generative,duan2017one} or learning from demonstration~\cite{SchaalNIPS1997}. Inverse reinforcement learning~\cite{Ng:2000:AIR:645529.657801,Abbeel:2004:ALV:1015330.1015430,wulfmeier2015maximum}, behavior cloning~\cite{6796843}, and curiosity-based exploration~\cite{pathakICMl17curiosity} are also examples of research in this field. Typically, in all these formulations, expert demonstrations are provided as input.

The majority of such prior work assumes that the demonstrations contain both states and actions $\{ (\bm{s^i_0}, \bm{a^i_0}), ... ,(\bm{s^i_t}, \bm{a^i_t})\}$ and that these can be used to solve the problem of having only a sparse reward or a complete lack thereof. However, there are many cases in real-world environments in which such detailed action information is not readily available. For example, a typical schoolteacher does not tell students the exact amount of force to apply to each of their fingers while they are learning how to write.

As such, in this work, as our primary contribution, we propose a reinforcement learning method in which the agent learns an internal predictive model that is trained on the external environment from state-only trajectories by expert demonstrations. This model is not trained on both the state and action pairs. Hence, during each RL step, this method estimates an expected reward value on the basis of the similarity between the actual and predicted state values by the internal model. Therefore, the agent must learn to reward known good states and penalize unknown deviations. Here, we formulate this internal model as a temporal-sequence prediction model that predicts the next state value given the current and past state values at every time step. This paper presents experimental results on multiple environments with varying input and output settings for the internal model. In particular, we show that it is possible to learn good policies using an internal model trained by observing only game-playing videos, akin to the way we as humans learn by observing others. Furthermore, we compare the performance of our proposed method regarding the baselines of \textit{hand-crafted} rewards, prior research efforts, and other \textit{baseline} methods for the different environments.

\section{Related Work}
In RL, an agent learns a policy $\pi(\bm{a}_t|\bm{s}_t)$ that produces good actions from the observation at the time. DQN~\cite{Mnih2015} showed that a Q-value function $q(s_t, a_t)$ can be successfully approximated with a deep neural network. DAQN~\cite{kimura2018daqn} showed the pre-training by a generative model reduces the number of training iterations. Similarly, actor and critic networks in DDPG can enable continuous control, e.g. in robotic manipulation by minimizing a distance between the robot end-effector and the target position. Since the success with DDPG, other methods, such as TRPO~\cite{icml2015_schulman15} and proximal policy optimization~(PPO)~\cite{schulman2017proximal} have been proposed as further improvements for model-free RL regarding continuous control.

Although RL enables an agent to learn an optimal policy in the absence of supervised training data, in a standard case, it involves the difficult task of \textit{hand-crafting} good reward functions for each environment~\cite{Abbeel:2004:ALV:1015330.1015430}. Several kinds of approach have been proposed to work around or tackle this problem. An approach that does not require \textit{hand-crafted} rewards is behavior cloning based on supervised learning instead of RL~\cite{6796843}. It learns the conditional distribution of actions from given states in a supervised manner. Although it has an advantage of fast convergence~\cite{duan2017one} (as behavior cloning learns a single action from states during each step), it typically results in the compounding of errors in future states.

An alternate approach, inverse reinforcement learning~(IRL), was proposed~\cite{Ng:2000:AIR:645529.657801}. In this work, the authors tried to recover the reward function as the best description of the given expert demonstrations from humans or expert agents using linear programming methods. This was based on the assumption that expert demonstrations are solutions to a Markov Decision Process~(MDP) defined by a hidden reward function~\cite{Ng:2000:AIR:645529.657801}. It demonstrated successful estimation of the reward function regarding relatively simple environments, such as a grid world and the mountain car problem. Extending \cite{Ng:2000:AIR:645529.657801}, entropy-based methods that compute a suitable reward function by maximizing the entropy of the expert demonstrations have been proposed~\cite{Ziebart200810030}. In another paper~\cite{Abbeel:2004:ALV:1015330.1015430}, a method was proposed for recovering the cost function on the basis of expected feature matching between observed policies and agent behavior. Furthermore, the research showed that it is necessary for the agent to imitate the behavior of the expert. Another use of the demonstrations is it was used for initializing the value function~\cite{Wiewiora:2003:PSQ:1622434.1622441}. 

Recently, there were some studies that extended such framework using deep networks as non-linear function approximators for both the policies and the reward functions~\cite{wulfmeier2015maximum}. In another relevant paper~\cite{ho2016generative}, the imitation learning problem was formulated as a two-player competitive game in which a discriminator network tries to distinguish between expert trajectories and agent-generated trajectories. The discriminator is used as a surrogate cost function which guides the agent's behavior to imitate the expert’s behavior by updating policy parameters on the basis of TRPO~\cite{icml2015_schulman15}. Recent related work also includes model-based imitation learning~\cite{pmlr-v70-baram17a} and robust imitation learning~\cite{DBLP:journals/corr/WangMRWFH17} using generative adversarial networks. It can be argued that our method is similar to the reward shaping method proposed by~\cite{Brys:2015:RLD:2832581.2832716} because both methods calculate the similarity of demonstrations as a reward shaping function. However, while their paper dealt only with discrete action tasks, we show a similar approach can be applied to continuous action tasks~\footnote{Please note they used a different Mario game from that used in this paper.}. Moreover, all the above-mentioned methods rely on both state and action information provided by expert demonstrations. 

Another recent line of work aimed at learning useful policies for agents even in the absence of expert demonstrations. In this regard, they trained an RL agent with a combination of intrinsic curiosity-based reward and hand-engineered reward that had a continuous or very sparse scalar signal~\cite{pathakICMl17curiosity}. The curiosity-based reward was designed to have a high value when the agent encountered unseen states and a low value when it was in a state similar to the previously explored states. The paper reported good policies in games, such as Super Mario Bros. and Doom, without any expert demonstrations. Here, we also compared our proposed method with the curiosity-based approach and demonstrated better-learned behavior. However, as a limitation, our method assumed that state demonstrations were available as expert data.

Also, there is a work that estimate the reward from linear function~\cite{Suay:2016:LDS:2936924.2936988}. However, they evaluated by simple task; specifically, they used 27 discrete state-variables for Mario. On the other hand, our method is using the non-linear model. 

At the same time as this work, a recent paper~\cite{bco} proposed learning policies using behavior cloning method based on observations only. Unlike that work, here we put primary focus on reward shaping based on internal model from observation data.

\section{Proposed Method}
\subsection{Problem Statement}
We considered a MDP consisting of states $\mathcal{S}$ and actions $\mathcal{A}$, where the reward signal $r : \mathcal{S} \times \mathcal{A} \rightarrow \mathbb{R}$ was unknown. An agent acted defined by this MDP following a policy, $\pi(\bm{a_t}|\bm{s_t})$. Here, we assumed to have knowledge of a finite set of expert state trajectories, $\tau = \{ \bm{S^0}, ...,\bm{S^n}\}$, where $S^i = \{ \bm{s^i_0}, ...,\bm{s^i_m}\} $. These trajectories represented joint angles, raw images, or other environmental states.

Since the reward signal was unknown, our primary goal was to find a reward signal that enabled the agent to learn a policy, $\pi$, that could maximize the likelihood of these sets of expert trajectories, $\tau$. In this paper, we assumed that the reward signal could be inferred entirely on the basis of the information of the current and following states, $r : \mathcal{S} \times \mathcal{S} \rightarrow \mathbb{R}$. More formally, we wanted to find a reward function that maximized the following objective:

\begin{equation}
{r^*} = \argmax _{r} {{}\mathbb{E}}_{p(\bm{s_{t\texttt{+}1}}|\bm{s_{t}})} r(\bm{s_{t\texttt{+}1}}|\bm{s_{t}})
\label{eq:rr},
\end{equation}

\noindent where $r(\bm{s_{t\texttt{+}1}}|\bm{s_{t}})$ is the reward function of the next state on the basis of the current state and $p(\bm{s_{t\texttt{+}1}}|\bm{s_{t}})$ is the transition probability. We hypothesized that maximizing the likelihood of the next step prediction in \eqq\ref{eq:rr} resulted in increasing future rewards. This is because the likelihood was based on the similarity of current state values with the demonstrations obtained using the expert agent, which inherently chooses actions that would maximize their expected future reward. As such, we assumed the agent maximized the reward when it took the action that changed to a similar step value with given states from the expert.

\subsection{Training the Internal Model}
\label{sec_internal}
Let $ \tau = \{ \bm{s}_t^i \}_{i=1:M, t=1:N} $ be the expert states obtained by the expert agent, where $M$ is the number of demonstration episodes and $N$ is the number of steps within each episode. We trained the internal model to predict reward signals on the basis of the expert state trajectories, $\tau$, which in turn were used to guide a reinforcement learning algorithm and learn a suitable policy.

A simple straightforward idea~(\textit{baseline}) for an internal model is to use a generative model of the state value,~$\bm{s_{t}^i}$, to understand the $\tau$. The model trains a distribution of the state values, from which a predicted reward can be estimated on the basis of a similarity between the reconstructed state value and the actual experienced state value. This method constrains exploration to the states that have been demonstrated by experts and enables learning a policy in a way that closely matches that of the expert. However, the temporal order of states is ignored or not readily accounted for, and the temporal order of the next state in the sequence is important for estimating the state transition probability function. 

Therefore, our proposed method uses a recurrent neural network~(RNN)-based temporal-sequence model as an internal model that can be trained to predict the next state value given current and previous states on the basis of the expert trajectories. Such RNN temporal-sequence prediction models have been used successfully in the past as internal forward models in the context of grammar learning and robot behavior prediction~\cite{bakker2002reinforcement,10.3389/fnbot.2015.00010}. Here, we trained a deep temporal sequence prediction model as the internal model by using the given state values,~$\bm{s_{t}^i}$, and the next state values,~$\bm{s_{t\texttt{+}1}^i}$, from the expert demonstration trajectories, $\tau$. The model was trained to maximize the likelihood of the next state, such that the objective function for the model was:

\begin{equation}
\bm{\theta^ *} = \argmin _{\bm{\theta}} \Big [-\sum_{i=1}^M \sum_{t=1}^N \log p(\bm{s_{t\texttt{+}1}^i}|\bm{s_{t}^i}; \bm{\theta}) \Big],
\end{equation}

\noindent where $\bm{\theta^*}$ represents the optimal parameters of the internal model. We also assumed the probability of the next state given the previous state value, $p(\bm{s^i_{t\texttt{+}1}}|\bm{s^i_{t}};\bm{\theta})$, to be a Gaussian distribution. As such, the objective function could be seen as minimizing the mean square error, $\|\bm{s^i_{t\texttt{+}1}} - \bm{\theta}(\bm{s^i_t}) \|_{2}$, between the actual next state, $\bm{s^i_{t\texttt{+}1}}$, and the predicted next state, $\bm{\theta}(\bm{s^i_t})$. 

\subsection{Reinforcement Learning}
During the reinforcement learning, the method predicts a reward value with the trained internal model. The value is estimated as a function of the similarity between an actual next state value,~$\bm{s_{t\texttt{+}1}}$, and the predicted next state value,~$\bm{\theta}(\bm{s_t})$, given the current state value,~$\bm{s_{t}}$. Thus the reward function is formulated as:

\begin{equation}
r_t = - \psi \Big ( \|\bm{s_{t\texttt{+}1}} - \bm{\theta}(\bm{s_t}) \|\Big ),
\end{equation}

\noindent where $\psi$ is a function that reshapes the reward structure. In this paper, we tried a normal linear function, a hyperbolic tangent function, and a Gaussian function as the $\psi$ function. In this formulation, if the current state was similar to the predicted state value, the estimated reward value was high. However, if the current state was not similar to the predicted state, the reward value was low. Moreover, as the reward value was estimated at each time step, this approach could predict dense rewards even regarding problems in which the original \textit{hand-crafted} reward had a sparse structure.

Algorithm~\ref{alg:proposed_method} explains the flow of the method. The RL procedure is shown as part of a generic RL pipeline and can be implemented with most on- or off-policy RL algorithms. In this paper, we used DDPG and DQN RL algorithms.

\begin{algorithm}[tb]
\caption{Reinforcement Learning with Internal Model}\label{euclid} 
\label{alg:proposed_method}
\begin{algorithmic}[1]
\Procedure{Training Demonstrations}{}
\State $ \text{Given }\textit{trajectories} \: \tau \text{ from expert agent}$
\For{$ \bm{s_{t}^i}, \bm{s_{t\texttt{+}1}^i} \in \tau$}
\State $ \bm{\theta^*} \gets \argmin_{\bm{\theta}} \Big [-\sum_{i,t}\log p(\bm{s_{t\texttt{+}1}^i}|\bm{s_{t}^i};\bm{\theta}) \Big] $
\EndFor{}
\EndProcedure
\Procedure{Reinforcement Learning}{}
\For{$t = 1, 2, 3, ...$}
\State $ \text{Observe }\textit{state} \: \bm{s_{t}}$
\State $ \text{Execute }\textit{action} \: \bm{a_t} \text{, and observe }\textit{state} \: \bm{s_{t\texttt{+}1}}$
\State $ r_t \gets - \psi \Big( \|\bm{s_{t\texttt{+}1}} - \bm{\theta}(\bm{s_t}) \| \Big)$
\State $ \text{Update network~\footnotemark~using } (\bm{s_t}, \bm{a_t}, r_t, \bm{s_{t\texttt{+}1}}) $
\EndFor{}
\EndProcedure
\end{algorithmic}
\end{algorithm}

\footnotetext{The parameter updates are carried out following the standard procedure of the specific reinforcement learning (on-policy or off-policy) algorithm.}

\section{Experiment}

We conducted experiments across a range of environments. We prepared four different tasks with varying complexity, namely, controlling a robot arm so that the end-effector reaches a target position, controlling a point agent to move to a target point while avoiding an obstacle, sending commands to a bird agent for the longest flight in the Flappy Bird video game, and controlling the Mario agent to maximize a total travelled distance in the Super Mario Bros video game. \tabb\ref{tab:comparison_env} summarizes the key differences between the experiments. 

\renewcommand{\arraystretch}{1.5}
\begin{table}[tb]
\begin{center}
\begin{tabular}{ccccc}
\hline
Environment & Input & Action & RL\\
\hline
Reacher & joint angle & continuous & DDPG \\
Mover w/ obstacle & pos., dist.\footnotemark & continuous & DDPG \\
Flappy Bird & image, pos. & discrete & DQN \\
Super Mario Bros. & image & discrete & A3C \\
\hline
\end{tabular}
\end{center}
\caption{Comparison of different environments.}
\label{tab:comparison_env}
\end{table}
\renewcommand{\arraystretch}{1.0}

\subsection{Reacher}

\begin{figure}[tb]
\begin{minipage}{0.35\linewidth}
\centering
\includegraphics[width=2.8cm]{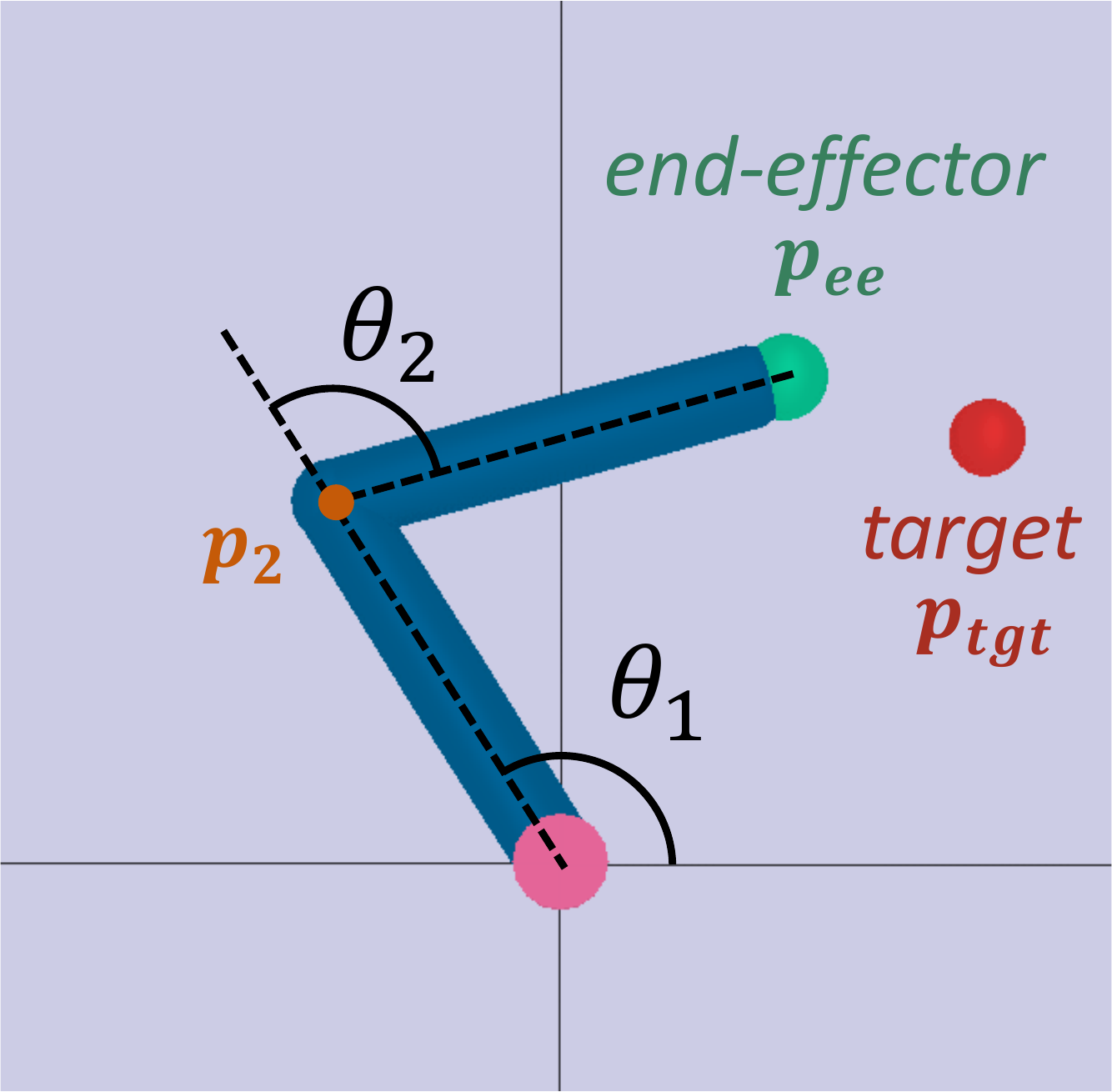}
\caption{Reacher environment. Objective of agent is to make end-effector~\textit{(green)} reach target~\textit{(red)}.}
\label{fig:reacher1_setting}
\end{minipage}
\hspace{10pt}
\begin{minipage}{0.55\linewidth}
\centering
\includegraphics[width=5cm]{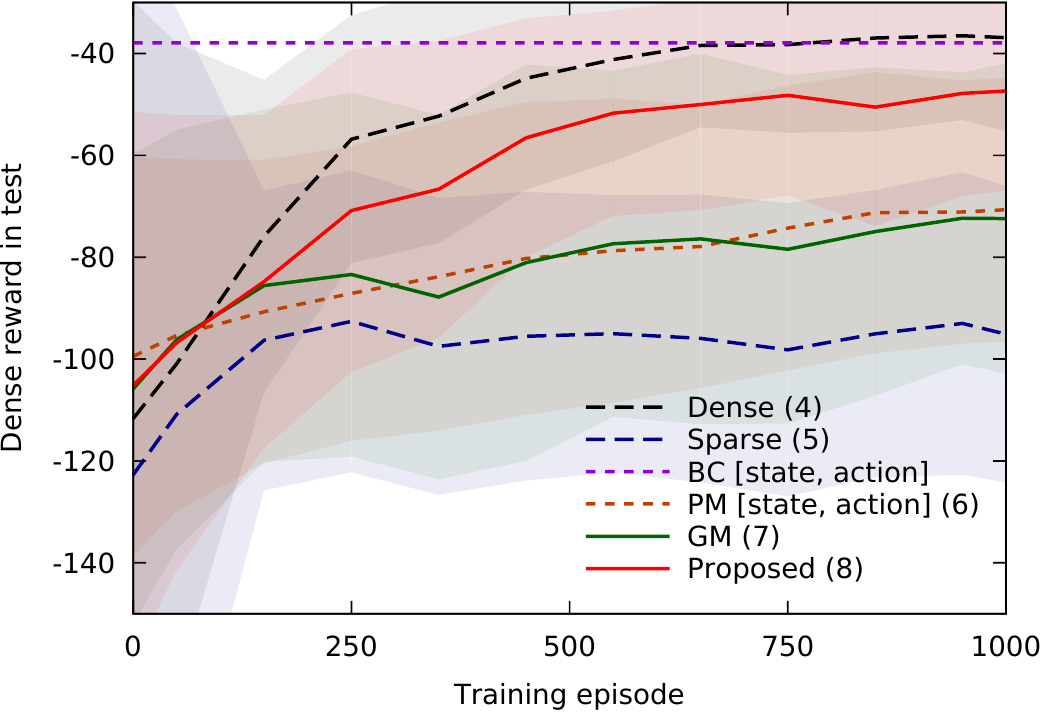}
\caption{Performance of RL for reacher. Number in brackets corresponds to equation number.}
\label{fig:reacher2_result}
\centering
\end{minipage}
\end{figure}

We considered a two degree of freedom (2-DoF) robot arm in an x-y plane that has to learn to make the end-effector reach a target position. The first link of the robot was rigidly connected to the $(0,0)$ point, and the second link connected to an edge of the first link. It had two joint values: $\bm{\theta} = (\theta_1, \theta_2), \: \theta_1 \in (-\infty,+\infty)$ and $\theta_2 \in [-\pi, +\pi]$, and the lengths of the links were $0.1$ and $0.11$, respectively. The~$\bm{p_2}$ is end point of the first link, and the~$\bm{p_{ee}}$ is the end-effector position of two links. The joint values and a target position were initialized by random values at the initial step of each episode. Specifically, the $x$ and $y$ of target position, $ \bm{p_{tgt}}$, were set from a random uniform distribution of $[-0.27, +0.27]$. The applied continuous action value, $\bm{a_t}$, was used to control the joint angles, such that $\dot{\bm{\theta}} = 0.05 \:\bm{a_t}$. Each action value was clipped within the range of $[-1, 1]$. The state vector, $\bm{s_t}$, consisted of the following variables: an absolute end position of the first link ($\bm{p_2}$), a joint value between the first link and the second link ($\theta_2$), velocities of the joints ($\dot{\theta}_1, \dot{\theta}_2$), and an absolute target position ($\bm{p_{tgt}}$). We used the roboschool environment with built-in physical dynamics~\cite{DBLP:journals/corr/BrockmanCPSSTZ16,roboschool} for this experiment. \figg\ref{fig:reacher1_setting} illustrates the used environment. The robot links are in blue, the green point is the end-effector, and the red point is the target location.

We used the DDPG algorithm~\cite{ddpg} to train the RL agent. The actor and critic-network had $400$, $300$ fully-connected~(FC) neuron layers, respectively. The output from the final layer of the actor was passed through a $\tanh$ activation function while others passed through the ReLU~\cite{icml2010_NairH10} activation function. The exploration policy was an Ornstein-Uhlenbeck process~\cite{PhysRev.36.823}, the size of replay memory was $1$ million steps, and we used the Adam optimizer~\cite{DBLP:journals/corr/KingmaB14} for the stochastic gradient updates. The number of steps for each episode was set to $400$ in this experiment. All implementations were done using the Keras-rl~\cite{plappert2016kerasrl} and Keras~\cite{chollet2015keras} libraries. Here, we compared the following reward functions:

\footnotetext{``pos.'' implies position, and ``dist.'' implies distance.}

\begin{flalign}
\hspace{10pt}\text{Hand}&\text{-crafted dense reward:} \notag\\
r_t &= - \|\bm{p_{ee}} - \bm{p_{tgt}} \|_{2} + r_t^{env}& \\
\hspace{10pt}\text{Hand}&\text{-crafted sparse reward:} \notag\\
r_t &= - 100 \tanh( \|\bm{p_{ee}} - \bm{p_{tgt}} \|_{2}) + r_t^{env}&\\
\hspace{10pt}\text{Predi}&\text{ctive model (PM, with state-action pair):} \notag\\
r_t &= - 10 \tanh( \|\bm{s_{t\texttt{+}1}} - \bm{\theta_{\texttt{+}a}}(\bm{s_{t}}, \bm{a_{t}}) \|_{2}) + r_t^{env} \\
\hspace{10pt}\text{Gene}&\text{rative model (GM, \textit{baseline}):} \notag\\
r_t &= - \tanh( \|\bm{s_{t\texttt{+}1}} - \bm{\theta_g}(\bm{s_{t\texttt{+}1}}) \|_{2}) + r_t^{env}& \\
\hspace{10pt}\text{\textbf{Prop}}&\text{\textbf{osed method}:} \notag\\
r_t &= - 10 \tanh( \|\bm{s_{t\texttt{+}1}} - \bm{\theta}(\bm{s_{t}}) \|_{2}) + r_t^{env}&
\end{flalign}
\noindent where $r_t^{env}$ is an environment specific reward, which is the cost for current action, $- \| \bm{a_t} \|_{2}$. This regularization was required to find the shortest path to reach the target. The expert demonstrations,~$\tau$, had $2000$ episode trajectories by running a trained agent. The model,~$\bm{\theta_{\texttt{+}a}}$, used both state-action pairs to estimate the reward function, $\|\bm{s_{t+1}^i} - \bm{\theta_{\texttt{+}a}}(\bm{s_t^i}, \bm{a_t^i}) \|_{2}$, where $\bm{s_t^i}$ and $\bm{a_t^i}$ were obtained from demonstrations. Our proposed internal model was not required such action information $\bm{a_t^i}$. The proposed method was constructed using long short-term memory~(LSTM)~\cite{Hochreiter:1997:LSM:1246443.1246450} as with the temporal sequence model. The model had two 128-unit LSTM layers with $\tanh$ activation and a $40$-unit FC layer with ReLU activation. Furthermore, we also compared it with a standard behavior cloning~(BC)~\cite{6796843} procedure, which used the actor-network directly trained with state-action pairs from expert demonstrations.

\figg\ref{fig:reacher2_result} shows the performance of the agents. In all cases, using internal-model-based rewards gave better results than having sparse rewards. Moreover, the model-based learning curves started from a better initial point compared to the dense reward curve. As observed, our proposed method achieved the best results when compared with all the baseline methods and also nearly achieved the results obtained in the dense reward case. As expected, the GM failed to work well in this complex experiment. The PM model with state-action information also performed poorly. However, in comparison, the BC method worked relatively well. This is not surprising and clearly indicates that it is better to use behavior cloning than reward prediction when both state and action information are available from expert demonstrations.

\subsection{Mover with Obstacle}
\begin{figure}[tb]
\begin{minipage}{0.3\linewidth}
\centering
\includegraphics[width=2.5cm]{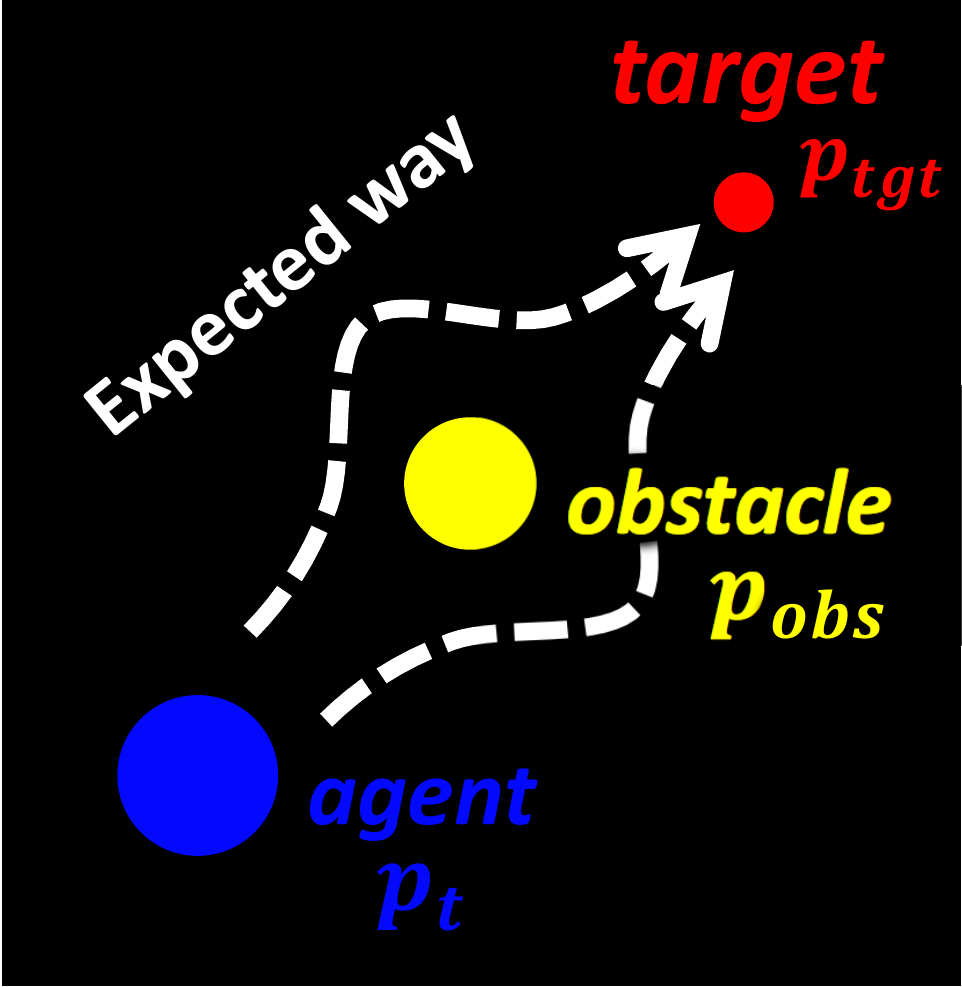}
\caption{Mover with obstacle. Objective of agent \textit{(blue)} is to move to target~\textit{(red)} while avoiding obstacle~\textit{(yellow)}.}
\label{fig:mover1_setting}
\end{minipage}\hfill
\begin{minipage}{0.65\linewidth}
\centering
\includegraphics[width=5.5cm]{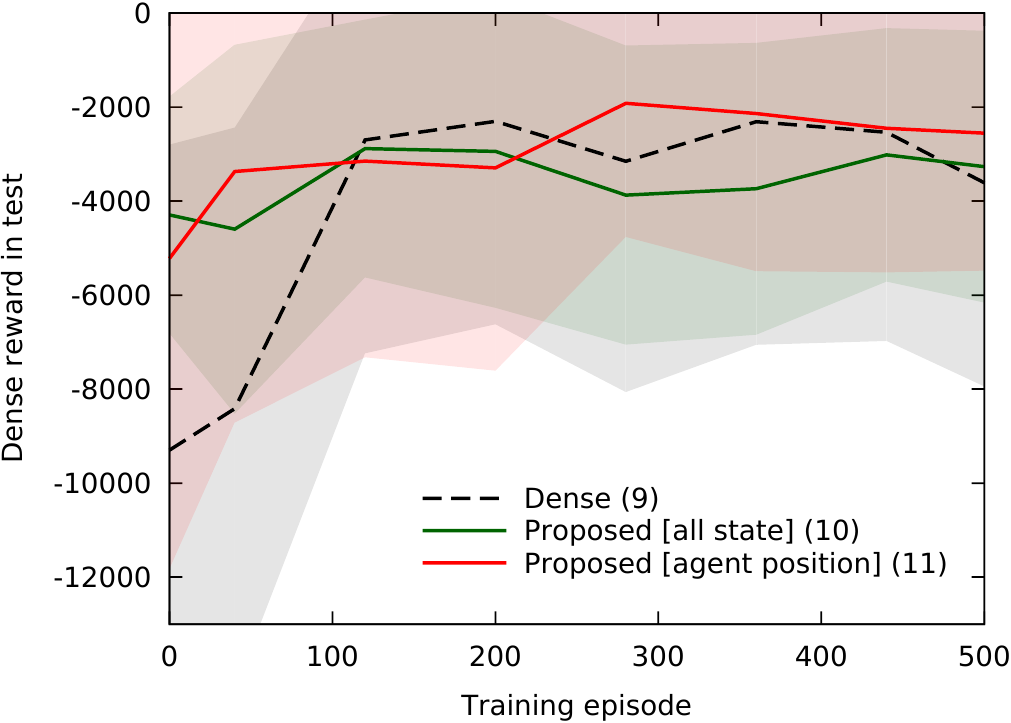}
\caption{Performance for mover with obstacle. We tried two different conditions for proposed method.}
\label{fig:mover1_result}
\end{minipage}
\end{figure}

In this task, we developed a new environment which has position control and an obstacle. The task was to move toward a target position without colliding with the obstacle. \figg\ref{fig:mover1_setting} illustrates the environment setup. The initial position of the agent, the target position, and the obstacle’s position were initialized randomly. The state vector, $\bm{s_t}$, contained the following variables: the agent’s absolute position~$(\bm{p_t})$, the current velocity of the agent~$(\bm{\dot{p}_t})$, the target position~$(\bm{p_{tgt}})$, the obstacle’s position~$(\bm{p_{obs}})$, and the relative target and obstacle location regarding the agent~$(\bm{p_t}-\bm{p_{tgt}}, \bm{p_t}-\bm{p_{obs}})$. The RL algorithm used was DDPG~\cite{ddpg}; the actor and critic networks had $64$ and $64$-unit FC layers, and each layer had a ReLU activation function. The exploration policy was the Ornstein-Uhlenbeck process~\cite{PhysRev.36.823}, the size of the replay memory was $500$ thousand, and the optimizer was Adam. The number of steps for each episode was set to 500. 

Here, we tried predicting a part of the state that is related to a given action, thus taking the relevance into account. In former work~\cite{pathakICMl17curiosity}, the authors predicted the function of the next state, $\phi(\bm{s_{t\texttt{+}1}})$, rather than predicting the actual value, $\bm{s_{t\texttt{+}1}}$. In this experiment, we chose the agent position,~$(\bm{p_t})$, as the selected state value. Furthermore, we changed the non-linear function, $\psi$, to a Gaussian function. This allowed us to compare the robustness of our proposed method when using different non-linear functions. Here, we used the following reward functions:
\begin{flalign}
\hspace{10pt}\text{Hand}&\text{-crafted dense reward:} \notag\\
r_t &= - \|\bm{p_t} - \bm{p_{tgt}} \|_{2} + \|\bm{p_t} - \bm{p_{obs}} \|_{2} &\\
\hspace{10pt}\text{\textbf{Prop}}&\text{\textbf{osed method} (predict next state values):} \notag\\
r_t &= \exp( - \|\bm{s_{t\texttt{+}1}} - \bm{\theta}(\bm{s_{t}}) \|_{2} / 2 \sigma_1^2 ) &\\ 
\hspace{10pt}\text{\textbf{Prop}}&\text{\textbf{osed method} (predict only next agent position):} \notag\\
r_t &= \exp( - \|\bm{s'_{t\texttt{+}1}} - \bm{\theta'}(\bm{s_{t}}) \|_{2} / 2 \sigma_2^2 ),&
\end{flalign}
\noindent where $\bm{s_{t}'}$ is the agent’s position, $\bm{\theta'}$ is an internal network that predicts a selected state~$\bm{\hat{s}_{t}'}$, $\sigma_1$ is 0.005, and $\sigma_2$ is 0.002. The dense reward was composed of both the target distance cost and an obstacle distance bonus. The expert trajectories, $\tau$, contained 800 \textit{human-guided} demonstration data with only state values; therefore, behavior cloning could not be directly applied. The internal prediction model once again used an LSTM network that consisted of two $256$-unit LSTM layers with ReLU activations.

\figg\ref{fig:mover1_result} shows the performance obtained with the different reward settings. As observed, the proposed internal model learned to reach the target faster than the dense reward. Using the agent’s position prediction internal model achieved the best performance. 

\subsection{Flappy Bird}

In this experiment, we used a re-implementation~\cite{Keras-FlappyBird} of the ``Flappy Bird'' game. The objective of this game is to make the agent pass through as many pipes as possible without collision. The control is a single discrete command of whether to flap the bird’s wings or not. The RL state value had four consecutive gray frames~(4~$\times$~80~$\times$~80~pixels). A well-trained agent can play for an arbitrary number of steps; however we limited 1000 steps for each episode. And the each position of the pipe is random.
In this case, we used the DQN~\cite{Mnih2015} RL algorithm in which the network had three convolutional and two FC layers. Each layer had ReLU activation, and it used the Adam optimizer and mean-squared loss. The size of replay memory was 2 million steps, the batch size was 256, and all other parameters were fixed following the original implementation~\cite{Keras-FlappyBird}. The update frequency of the deep network was 100 steps. Here, we compared the following rewards:

\begin{flalign}
\hspace{10pt}\text{Hand}&\text{-crafted reward~(the point for game):} \notag\\
r_t &= \begin{cases}
 +0.1 & \text{if \textit{ alive}} \\
 +1 & \text{if \textit{ passes through a pipe}} \\
 -1 & \text{if \textit{ collides with a pipe}}
\end{cases} &\\
\hspace{10pt}\text{\textbf{Prop}}&\text{\textbf{osed method} (predict next bird position):} \notag\\
r_t &= \exp( - \|\bm{s'_{t\texttt{+}1}} - \bm{\theta'}(\bm{s_{t}}) \|_{2} / 2 \sigma^2 )&,
\end{flalign}
\noindent where $\bm{s'_{t}}$ is the absolute position of the bird that can be given from the simulator, and $\sigma$ is 0.02. The absolute position was not in the state value; however, it can be estimated by simple image processing. The internal model, $\bm{\theta'}$, was constructed using an LSTM network to predict the bird’s next position given the image input. The set of expert trajectories, $\tau$, had only $10$ episodes obtained from a trained agent available from the Github repository~\cite{Keras-FlappyBird}. In this case, we also compared the learned agent behavior with that obtained using a behavior cloning method.

\begin{figure}[tb]
\begin{minipage}{0.44\linewidth}
\centering
\includegraphics[width=3.6cm]{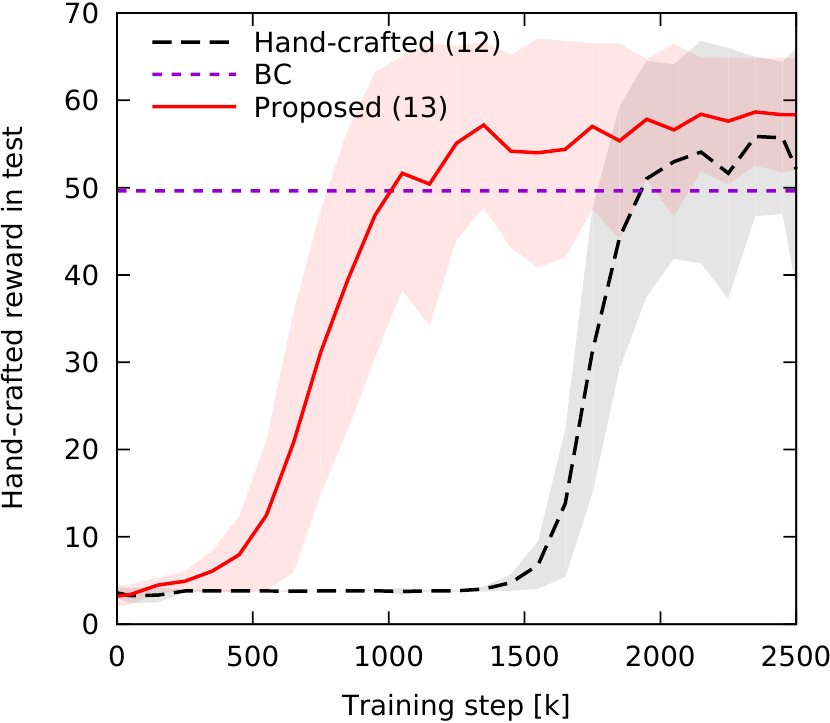}
\caption{Performance for Flappy Bird~(k is $10^3$). Proposed method trains 10 episodes.}
\label{fig:flappy_result}
\end{minipage}\hfill
\begin{minipage}{0.52\linewidth}
\centering
\includegraphics[width=4.5cm]{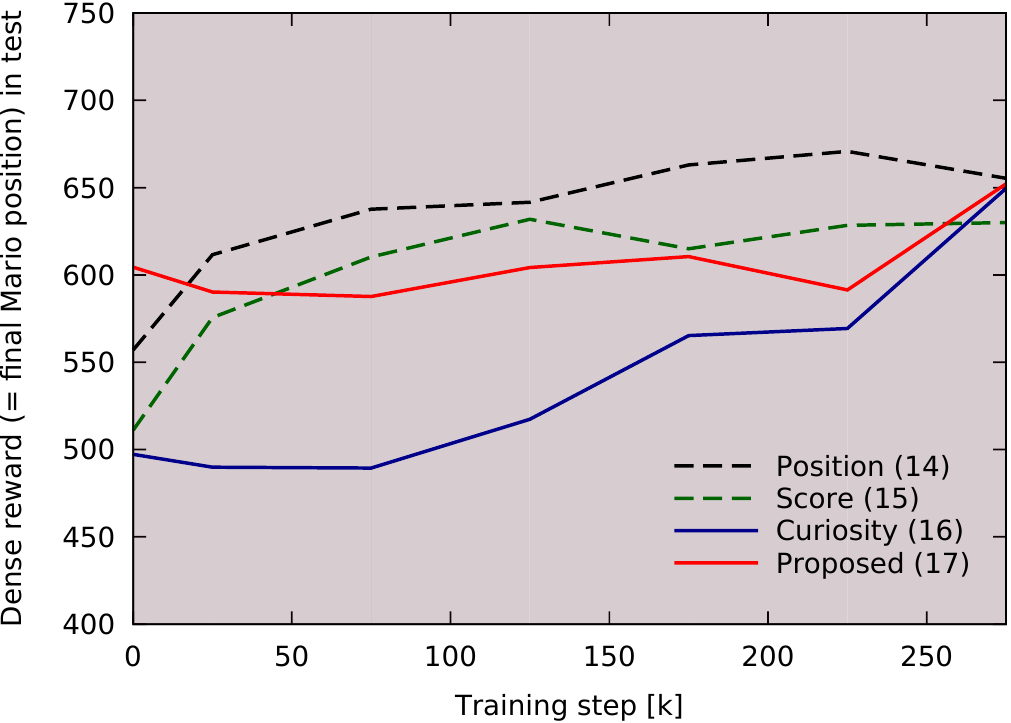}
\caption{Performance for Super Mario Bros. Proposed method trains only 15 videos without any meta data.}
\label{fig:mario_result}
\end{minipage}
\end{figure}

\figg\ref{fig:flappy_result} clearly demonstrates that our proposed method converges faster than \textit{hand-crafted} rewards. This can be ascribed to the fact that the hand-crafted reward only took into account the distance traveled, whereas, our internal model estimated reward provides information about which absolute transitions are good. The hand-crafted reward of this game was the big positive value when it passed the pipe, otherwise it was small positive value when the bird was alive. This means the big positive value will be delayed even if the bird chose the good action. Even though it is given each step, the hand-crafted reward does not contain the detailed reward value for each transition. On the other hand, our method could estimate the detailed reward by using the similarity of state information for each transition. Furthermore, our proposed method converges significantly better with fewer demonstrations than the baseline BC method; the reason is the number of demonstration was small.

\subsection{Super Mario Bros.}

In the final task, we considered a more difficult setting so that we could obtain only raw state information to clarify the benefits of the proposed method. Here, we applied our internal model-based reward estimator to Nintendo's ``Super Mario Bros.'' game and used a classic Nintendo video game emulator~\cite{mario} for the environment. In this experiment, we compared our method with a curiosity-based method~\cite{pathakICMl17curiosity} using their implementation~\cite{noreward-rl}. However, we slightly modified the game implementation to always initialize Mario at the starting position rather than at a previously saved checkpoint. The game has a discrete control where an agent~(Mario) can make 14 types of action; however, a single action was repeated for six consecutive frames. The state, $\bm{s_{t}}$, consisted of sequential input of four 42 x 42-pixel gray-frame images with skipping every six frames. We used the A3C~\cite{mnih2016asynchronous} on-policy RL algorithm to evaluate our model. Moreover, we tried the gameplay of stage ``1-1'' of the game in this experiment. The main objective of the agent was to travel as far as possible. We compared the following rewards:
\begin{flalign}
\hspace{10pt}\text{Differ}&\text{ence of Mario's position~(dense reward):} \notag\\
r_t &= position_{t} - position_{t-1} &\\
\hspace{10pt}\text{Differ}&\text{ence of score~(sparse reward):} \notag\\
r_t &= score_{t} - score_{t-1} &\\
\hspace{10pt}\text{Curio}&\text{sity~\cite{pathakICMl17curiosity}:} \notag\\
r_t &= \eta \| \phi(\bm{s_{t\texttt{+}1}}) - \bm{\theta_F}( \phi(\bm{s_{t}}), \bm{a_{t}} ) \|_{2} &\\
\hspace{10pt}\text{\textbf{Propo}}&\text{\textbf{sed method} (predict next frame):} \notag\\
r_t &= max (0, - \| \bm{s'_{t\texttt{+}1}} - \bm{\theta'}(\bm{s_{t}}) \| + \zeta)&
\end{flalign}
\noindent where $position_{t}$ is Mario's current position value, $score_{t}$ is a score value, $\bm{s'_{t}}$ is the latest frame in $\bm{s_{t}}$, and $\zeta$ is $0.025$. Position, score, and related meta-information could be directly obtained from the emulator. In our proposed method, we took $15$ game playing videos, each showing a single episode, from five different expert players and provided the demonstration trajectories, $\tau$. In total, $\tau$ consisted of $25$ thousand frames without any action or meta-information. We skipped $36$ frames to generate $\bm{s_t^i}$ because people cannot play as fast as an RL agent. We used a three-dimensional convolutional neural network~(3D-CNN)~\cite{ji20133d} as the $\bm{\theta'}$ model. The internal model, $\bm{\theta'}$, predicted the next frame image given the continuous frames, $\bm{s_t}$. The 3D-CNN network consisted of four convolutional layers\footnote{Two layers had (2 x 5 x 5) kernels, and the next two layers had (2 x 3 x 3) kernels. The all had 32 filters and (2, 1, 1) stride in every two layers.} and one final convolutional layer to reconstruct the image. Once again, the proposed method required only videos to train the internal model.

Here, we changed the $\psi$ function to a linear function to evaluate a simple formulation of the proposed method. However, a na\"ive reward estimate, ($r_t = - \|\bm{s'_{t\texttt{+}1}} - \bm{\theta'}(\bm{s_{t}}) \|_{2} \texttt{+} 1$),\footnote{The $\texttt{+}1$ reward was for the terminal condition.} does not work for this stage of the game. The Mario with the na\"ive method ends up getting positive rewards even if the agent remains stationary at the initial position (since enemy agents do not appear if Mario does not move). Hence, we applied a threshold, $\zeta$, value to prevent this trivial sub-optimal outcome. $\zeta$ was calculated on the basis of the reward value obtained by staying stationary at the initial position. 

\figg\ref{fig:mario_result} shows the performance with the different reward functions. The graph shows the mean learning curves across trials. As observed, the agent does not reach the goal every time, even with the \textit{hand-crafted} dense rewards~\footnote{The average position was 650, even with very long training steps, e.g. 3 million steps.}. This behavior was also observed in the original paper for their reward case~\cite{pathakICMl17curiosity}. However, as observed in \figg\ref{fig:mario_result}, our proposed method learns relatively faster than the curiosity- and score-based reward methods. Moreover, it was faster to obtain a good policy with the proposed method than with cases using dense rewards. 

Comparing with the flappy bird experiment, the position reward is representing about the goodness for each transition which means it is `dense' reward; the hand-crafted reward in the flappy bird was the delayed reward. We summarize the proposed method could generate the predicted dense reward, which is better value than sparse reward and has potential to become similar to dense reward, without any reward information. Also, this proposed reward helps a RL with reward as the reward shaping method. 

Regarding future work for Mario experiment, we believe using deeper networks as function approximators and high-resolution input images may improve the performance of the convergence further.

\section{Conclusion}
In this paper, we proposed a reinforcement learning method that uses an internal model based on expert-demonstrated state trajectories to predict rewards. This method does not require learning the dynamics of the external environment from state-action pairs. The internal model consisted of a temporal sequence predictive RNN for the given expert state distribution. During RL, the agent calculated the similarity between actual and predicted states, and this value was used to predict the reward. We compared our proposed methods with \textit{hand-crafted} rewards and previous methods in four different environments. Overall, we demonstrated that using internal model agents enables the learning of good policies, learning curves have better initialization, and learning converges faster than \textit{hand-crafted} reward and sparse reward in most cases. It was also shown that the method could be applied to cases in which the demonstration was obtained directly from videos by person. 

However, detailed trends were different for the different environments depending on the complexity of the task. As a current limitation of the method, we found that none of the rewards based on our proposed method were versatile enough to be applicable to every environment without any changes in the reward definition. There is room for further improvement, especially regarding modeling the global temporal characteristics of state trajectories. We would like to tackle the problem of generalizing across tasks in future work.

\bibliographystyle{ACM-Reference-Format}
\bibliography{ms}

\end{document}